\documentclass{article}
\usepackage{ijcai24}

\usepackage{times}
\usepackage{soul}
\usepackage{url}
\usepackage[hidelinks]{hyperref}
\usepackage[utf8]{inputenc}
\usepackage[small]{caption}
\usepackage{graphicx}
\usepackage{amsmath}
\usepackage{amsthm}
\usepackage{booktabs}
\usepackage{algorithmic}
\usepackage[ruled]{algorithm2e}
\usepackage[switch]{lineno}
\usepackage{svg}
\usepackage{color}
\usepackage{xcolor}
\usepackage{colortbl}
\usepackage{mathrsfs}
\usepackage{mathtools}
\usepackage{booktabs}

\usepackage{CJKutf8}
\usepackage{multirow}
\usepackage{enumitem}
\usepackage{svg}
\usepackage{amssymb}
\usepackage{pifont}
\usepackage{makecell}
\usepackage{array} 

\definecolor{mytablecolor}{HTML}{719EEE}
\definecolor{mytablecolor1}{HTML}{E5F3F9}
\definecolor{mytablecolor2}{HTML}{EFF4FD}

\title{Boosting Model Resilience via Implicit Adversarial Data Augmentation}

\author{
Xiaoling Zhou$^1$
\and
Wei Ye$^1$\footnotemark[1]\and
Zhemg Lee$^{2}$\and
Rui Xie$^1$\And
Shikun Zhang$^1$\footnotemark[1]\\
\affiliations
$^1$National Engineering Research Center for Software Engineering, Peking University, China\\
$^2$Tianjin University, Tianjin, China\\
\emails
xiaolingzhou@stu.pku.edu.cn,
wye@pku.edu.cn,
zhemglee@tju.edu.cn,
\{ruixie,zhangsk\}@pku.edu.cn
}

\begin{document}

\maketitle
\renewcommand{\thefootnote}{\fnsymbol{footnote}}
\footnotetext[1]{Corresponding authors.}
\renewcommand{\thefootnote}{\arabic{footnote}}

\begin{abstract}

Data augmentation plays a pivotal role in enhancing and diversifying training data. Nonetheless, consistently improving model performance in varied learning scenarios, especially those with inherent data biases, remains challenging. To address this, we propose to augment the deep features of samples by incorporating their adversarial and anti-adversarial perturbation distributions, enabling adaptive adjustment in the learning difficulty tailored to each sample’s specific characteristics. We then theoretically reveal that our augmentation process approximates the optimization of a surrogate loss function as the number of augmented copies increases indefinitely. This insight leads us to develop a meta-learning-based framework for optimizing classifiers with this novel loss, introducing the effects of augmentation while bypassing the explicit augmentation process. We conduct extensive experiments across four common biased learning scenarios: long-tail learning, generalized long-tail learning, noisy label learning, and subpopulation shift learning. The empirical results demonstrate that our method consistently achieves state-of-the-art performance, highlighting its broad adaptability.

\end{abstract}

\section{Introduction}
Data augmentation techniques, designed to enrich the quantity and diversity of training samples, have demonstrated their effectiveness in improving the performance of deep neural networks (DNNs)~\cite{maharana2022review}. Existing methods can be divided into two categories. The first, explicit data augmentation~\cite{r71,ijcai2023p177}, applies geometric transformations to samples and directly integrates augmented instances into the training process, leading to reduced training efficiency. The second, a recent addition to this field, is the implicit data augmentation technique~\cite{r8}, which is inspired by the existence of numerous semantic vectors within the deep feature space of DNNs. This method emphasizes enhancing the deep features of samples and is achieved by optimizing robust losses instead of explicitly conducting the augmentation process, resulting in a more efficient and effective approach. Subsequent studies in imbalanced learning have extended this approach.  
For example, 
MetaSAug~\cite{r9} refines the accuracy of covariance matrices for minor classes by minimizing losses on a balanced validation set. RISDA~\cite{r79} generates diverse instances for minor classes by extracting semantic vectors from the deep feature space of both the current class and analogous classes.

\begin{figure}[t] 
\centering
 \includegraphics[width=0.48\textwidth]{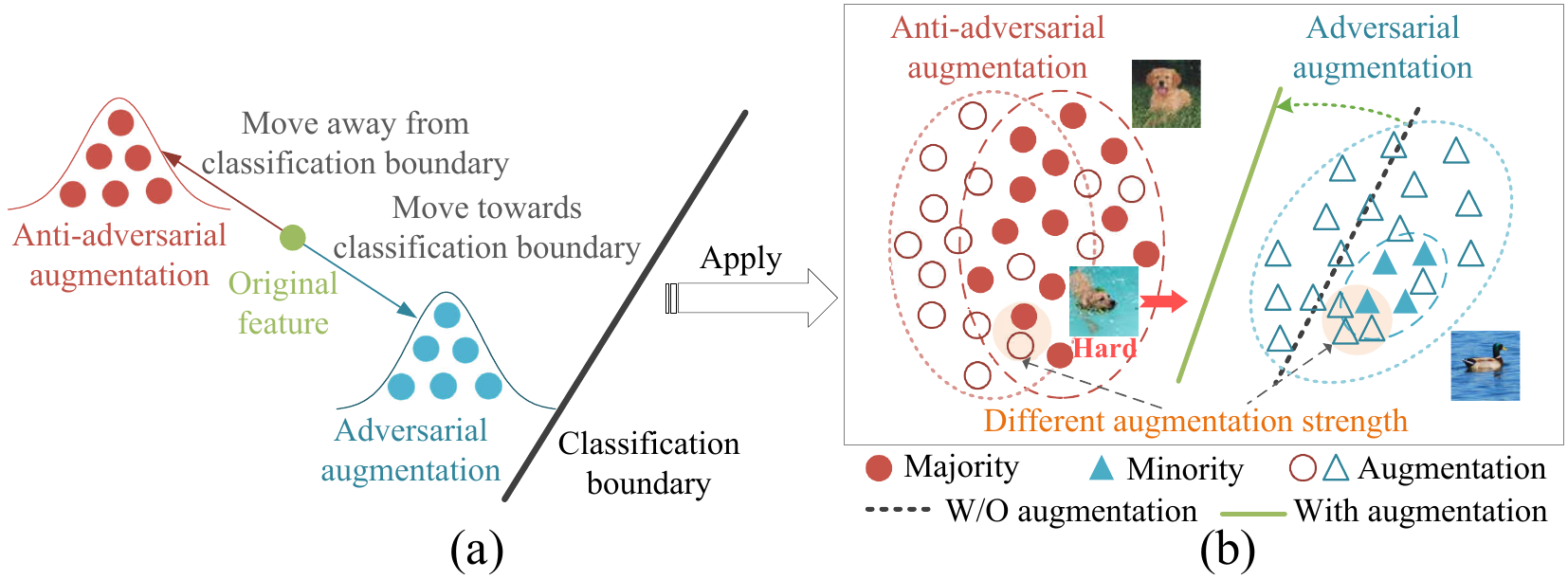}
\vspace{-0.22in}
\caption{
(a) Illustration for our augmentation strategy, which augments samples within their adversarial and anti-adversarial perturbation distributions.
(b) Illustration of an imbalanced learning scenario. Our method employs adversarial and anti-adversarial augmentations for the minor and major classes, respectively. 
}
\vspace{-0.05in}
\label{fig1}
\end{figure}

Despite these promising efforts, current augmentation strategies still exhibit notable limitations. Firstly, these approaches mainly enhance samples within the original training data space~\cite{ijcai2023p177,r8}, falling short in effectively mitigating the distributional discrepancies between training and test data, such as noise~\cite{zheng2021meta} and subpopulation shifts~\cite{r57}.
Secondly, previous algorithms~\cite{r9,r79} mostly operate at the category level, causing different samples within the same class to share identical augmentation distributions and strengths, which is potentially unreasonable and inaccurate. 
For instance, in noisy learning scenarios, it is advisable to treat noisy samples separately to mitigate their negative impact on model training~\cite{Zhou_Yang_Wu_2023}.

This study introduces a novel \textbf{I}mplicit \textbf{A}dversarial \textbf{D}ata \textbf{A}ugmentation (IADA)\footnote{Our code is available at \url{https://github.com/xiaolingzhou98/IADA}.} approach, which conceptually embodies two main characteristics to address the two bottlenecks above. Firstly, as illustrated in Fig.~\ref{fig1}(a), IADA enriches the deep features of samples by randomly sampling perturbation vectors from their adversarial and anti-adversarial perturbation distributions, surpassing the limitation of augmenting within the original training distribution. With this strategy, the classifier is anticipated to be dynamically adjusted by modifying the learning difficulty of samples. 
Secondly, the augmentation distribution for each sample is tailored based on its unique training characteristics, granting it the ability to address data biases beyond the category level. Specifically, these distributions are modeled as multivariate normal distributions, characterized by sample-wise perturbations and class-specific covariance matrices. 
Fig.~\ref{fig1}(b) demonstrates the two features of our method using imbalanced learning as an example scenario, where the classifier typically favors major classes while underperforming on the minor ones, as well as favors easy samples (e.g.,  dogs on grass) but commonly mispredicting hard samples (e.g., dogs in the water)\footnote{This example stems from the observation that the majority of dogs in the training set are on grass and seldom in the water.}. In this scenario, our method adopts anti-adversarial augmentation for major classes and adversarial augmentation (with higher strength) for minor ones. Meanwhile, adversarial augmentation is also applied to hard samples within major classes, ultimately facilitating better learning of class boundaries.

By exploring an infinite number of augmentations, we theoretically derive a surrogate loss for our augmentation strategy, thereby eliminating the need for explicit augmentation. Subsequently, to determine the perturbation strategies for samples in this loss, we construct a meta-learning-based framework, in which a perturbation network is tasked with computing perturbation strategies by leveraging diverse training characteristics extracted from the classifier as inputs. The training of all parameters in the perturbation network is guided by a small, unbiased meta dataset, enabling the generation of well-founded perturbation strategies for samples. 
Extensive experiments show that our method adeptly addresses various data biases, such as noise, imbalance, and subpopulation shifts, consistently achieving state-of-the-art (SOTA) performance among all compared methods.

In summary, our contributions in this paper are threefold.
\begin{itemize}[itemsep=1pt, topsep = 0pt,partopsep=0pt]
    \item 
We propose a novel perspective on data augmentation, wherein samples undergo augmentation within their adversarial and anti-adversarial perturbation distributions, to facilitate model training across diversified learning scenarios, particularly those with data biases. 
    \item 
In accordance with our augmentation perspective, we derive a new logit-adjusted loss and incorporate it into a well-designed meta-learning framework to optimize the classifier, unlocking the potential of data augmentation without an explicit augmentation procedure. 

    \item We conduct extensive experiments across four typical biased learning scenarios, encompassing long-tail (LT) learning, generalized long-tail (GLT) learning, noisy label learning, and subpopulation shift learning. The results conclusively demonstrate the effectiveness and broad applicability of our approach.
\end{itemize}

\section{Related Work}

\paragraph{Data Augmentation} methods have showcased their capacity to improve DNNs' performance by expanding and diversifying training data~\cite{maharana2022review}. Explicit augmentation directly incorporates augmented data into the training process, albeit at the expense of reduced training efficiency~\cite{r71,taylor2018improving,ijcai2023p177}. Recently, Wang et al.~\shortcite{r8} introduced an implicit semantic data augmentation approach, named ISDA, which transforms the deep features of samples within the semantic space of DNNs and boils down to the optimization of a robust loss. Subsequent studies~\cite{r9,r79} in image classification tasks have extended this approach. However, these methods still struggle with effectively improving model performance when dealing with data biases that go beyond the category level.

\begin{figure*}[t] 
\centering
\includegraphics[width=1\textwidth]{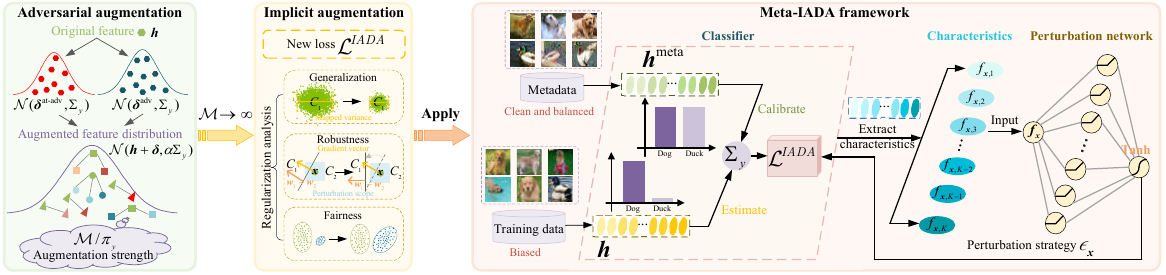}
\caption{The overview of our method pipeline. We initiate with a sample-wise adversarial data augmentation strategy (Box~1), enriching the deep features of samples using perturbation vectors extracted from their adversarial and anti-adversarial perturbation distributions. Subsequently, by considering an infinite number of augmented instances, we derive a novel robust loss, termed IADA (Box~2). Regularization analysis reveals the efficacy of IADA in improving model generalization, robustness, and inter-class fairness. To facilitate optimization with IADA, we then establish a meta-learning-based framework called Meta-IADA (Box~3). Within it, a perturbation network is tasked with generating perturbation strategies for samples (denoted as $\epsilon_{\boldsymbol{x}}$) in the IADA loss, leveraging a set of ($K\!=\!15$) training characteristics as inputs. 
}
\label{frametu}
\end{figure*}


\paragraph{Adversarial and Anti-Adversarial Perturbations} transform samples in directions that respectively move towards and away from the decision boundary, thereby modifying samples' learning difficulty~\cite{advtpami,Zhou_Yang_Wu_2023}. Consequently, models allocate varying levels of attention to samples subjected to their perturbations. Research has confirmed that incorporating adversarial and anti-adversarial samples during training assists models in achieving a better tradeoff between robustness and generalization~\cite{Zhou_Yang_Wu_2023,zhu2021understanding}. However, existing adversarial training methods primarily focus on two specific types of perturbations that maximize and minimize losses~\cite{xu2021robust,Zhou_Yang_Wu_2023}, posing limitations. Moreover, generating adversarial perturbations within the input space is time-consuming~\cite{madry2018towards}. Different from prior studies, our approach randomly selects perturbation vectors from both adversarial and anti-adversarial perturbation distributions, enabling the generation of multiple distinct adversarial and anti-adversarial samples. Furthermore, the perturbations are generated within the deep feature space, enhancing efficiency and ensuring universality across various data types.

\section{Implicit Adversarial Data Augmentation}

We initially introduce a sample-wise adversarial data augmentation strategy to facilitate model training across various learning scenarios. By considering infinite augmentations, we then derive a surrogate loss for our augmentation strategy. 



\subsection{Adversarial Data Augmentation}

Consider training a deep classifier $\mathcal{F}$, with weights $\boldsymbol{\Phi}$ on a training set, denoted as $\mathcal{D}^{{tr}} = \{(\boldsymbol{x}_{i}, y_{i})\}_{i=1}^{N}$, where $N$ refers to the number of training samples, and $y_{i} \in \{{1},\cdots,{\mathcal{C}}\}$ represents the label of sample $\boldsymbol{x}_{i}$. The deep feature (before logit) learned by $\mathcal{F}$ for $\boldsymbol{x}_{i}$ is represented as a $\mathcal{H}$-dimensional vector $\boldsymbol{h}_{i} = {\mathcal{F}}_{\boldsymbol{\Phi}}(\boldsymbol{x}_{i})\in \mathbb{R}^{\mathcal{H}}$.

Our augmentation strategy enhances samples within the deep feature space of DNNs. The perturbation vectors for the deep feature of each sample are randomly extracted from either its adversarial or anti-adversarial perturbation distributions. These distributions are modeled as multivariate normal distributions, $\mathcal{N}(\boldsymbol{\delta}_{i}, {\boldsymbol{\Sigma}}_{y_i})$, where $\boldsymbol{\delta}_{i}$ refers to the sample perturbation, and ${\boldsymbol{\Sigma}}_{y_i}$ represents the class-specific covariance matrix estimated from the features of all training samples in class $y_i$. As samples undergo augmentation within the deep feature space,  perturbations should also be generated within this space, facilitating semantic alterations for training samples. Consequently, the perturbation vector $\boldsymbol{\delta}_{i}$ for sample $\boldsymbol{x}_{i}$ is calculated as $\epsilon_{i}\!\cdot\!{sign}(\nabla_ {\boldsymbol{h}_{i}}\ell_{i}^{CE})$, where ${sign}(\nabla_{\boldsymbol{h}_{i}}\ell_{i}^{CE})$ signifies the gradient sign of the CE loss $\ell_{i}^{CE}$ with respect to $\boldsymbol{h}_{i}$. The parameter $\epsilon_{i}$ plays a pivotal role in determining the perturbation strategy applied to $\boldsymbol{x}_{i}$, encompassing both the perturbation direction and bound. Its positive or negative sign signifies adversarial or anti-adversarial perturbations, respectively. Furthermore, the absolute value $|\epsilon_{i}|$ governs the perturbation bound. In practical applications, the value of $\epsilon_{i}$ is dynamically computed through a perturbation network based on the training characteristics of $\boldsymbol{x}_{i}$, which will be elaborated in Section~\ref{frame}. Additionally, the class-specific covariance matrix $\boldsymbol{\Sigma}_{y_i}$ within this distribution aids in preserving the covariance structure of each class. Its value is estimated in real-time by aggregating statistics from all mini-batches, as detailed in Section~I of the Appendix. Regarding the augmentation strength quantified by the number of augmented instances $\mathcal{M}_{i}$ for $\boldsymbol{x}_{i}$, we define $\mathcal{M}_{i}$ as $\mathcal{M}/\pi_{y_{i}}$, where $\mathcal{M}$ is a constant and $\pi_{y_{i}}$ represents the proportion of class $y_i$ in the training data. Accordingly, a smaller proportion results in a larger number of augmented instances, ensuring class balance.

To compute the augmented features $\tilde{\boldsymbol{h}}_{i}$ from ${\boldsymbol{h}}_{i}$, we transform ${\boldsymbol{h}}_{i}$ along random directions sampled from 
$\mathcal{N}(\boldsymbol{\delta}_{i},{\boldsymbol{\Sigma}}_{y_i})$. This transformation yields $\tilde{\boldsymbol{h}}_{i}\sim\mathcal{N}({\boldsymbol{h}}_{i}+\boldsymbol{\delta}_{i},\alpha{\boldsymbol{\Sigma}}_{y_{i}})$, where the parameter $\alpha$ controls the extent of dispersion for augmented samples.
In summary, our adversarial data augmentation strategy offers the following advantages:

\begin{itemize}[itemsep=1pt, topsep=1pt]
 \item Instead of augmenting samples within the original data space, our approach enhances them within their adversarial and anti-adversarial perturbation distributions. This method effectively adjusts the learning difficulty distribution of training samples, 
    fostering improved generalization and robustness in DNNs. 
\item Our sample-wise augmentation distribution customizes the mean vector based on the unique training characteristics of each sample. This personalized strategy significantly enhances models' ability to address data biases, encompassing those beyond the category level.
   

   
\end{itemize}

\subsection{IADA Loss}

With our augmentation strategy, a straightforward way to train a classifier involves augmenting each ${\boldsymbol{h}}_{i}$ for $\mathcal{M}_{i}$ times. This procedure generates an augmented feature set for each sample, $\{\tilde{\boldsymbol{h}}_{i}^{1},\cdots, \tilde{\boldsymbol{h}}_{i}^{\mathcal{M}_{i}}\}$. Subsequently, the CE loss for all augmented samples is as follows:
\begin{equation}
\begin{aligned}
\mathcal{L}^{\mathcal{M}}(\boldsymbol{W},\boldsymbol{b})=\frac{1}{\hat{\mathcal{M}}}\sum_{i=1}^{N}\sum_{k=1}^{\mathcal{M}_{i}}-\log\frac{e^{{\boldsymbol{w}_{y_{i}}^{T}\tilde{\boldsymbol{h}}_{i}^{k}+b_{y_i}}}}{\sum_{j=1}^{\mathcal{C}}e^{{\boldsymbol{w}_{j}^{T}\tilde{\boldsymbol{h}}_{i}^{k}+b_{j}}}},
\end{aligned}
\end{equation}
where $\hat{\mathcal{M}}\!=\!\sum_{i=1}^{N}\mathcal{M}_{i}$. Additionally, $\boldsymbol{W}\!=\![\boldsymbol{w}_{1},\!\cdots\!, \boldsymbol{w}_{{\mathcal{C}}}]^{T}\!\in\! \mathbb{R}^{\mathcal{C}\times\mathcal{H}}$ and $\boldsymbol{b}=[b_{1},\cdots, b_{{{\mathcal{C}}}}]^{T}\in \mathbb{R}^{\mathcal{C}}$, in which $\boldsymbol{w}_{j}$ and $b_{j}$ refer to the weight vector and bias corresponding to the last fully connected layer for class $j$. Considering augmenting more data while enhancing training efficiency, we explore augmenting an infinite number of times for the deep feature of each training sample. As $\mathcal{M}$ in $\mathcal{M}_{i}$ approaches infinity, the expected CE loss is expressed as:
\begin{equation}
\begin{aligned}
\mathcal{L}^{\mathcal{M}\rightarrow\infty}(\boldsymbol{W},\boldsymbol{b})
    &=\sum_{i=1}^{N}\frac{1}{\pi_{y_i}}\mathrm{E}_{\tilde{\boldsymbol{h}}_{i}}\big[-
    \log\frac{e^{{\boldsymbol{w}_{y_{i}}^{T}\tilde{\boldsymbol{h}}_{i}+b_{y_i}}}}{\sum_{j=1}^{\mathcal{C}}e^{{\boldsymbol{w}_{j}^{T}\tilde{\boldsymbol{h}}_{i}+b_{j}}}}\big].
    \end{aligned}
    \label{kaishieq2}
\end{equation}
However, accurately calculating Eq.~(\ref{kaishieq2}) poses a challenge. Hence, we proceed to derive a computationally efficient surrogate loss for it. Given the concavity of the logarithmic function $\log\left(\cdot\right)$, using Jensen's inequality, $\mathrm{E}[\log X]\le \log \mathrm{E}[X]$, we derive an upper bound of Eq.~(\ref{kaishieq2}) as follows:
\begin{equation}
\resizebox{1\hsize}{!}{$
\begin{aligned}
    \mathcal{L}^{\mathcal{M}\rightarrow\infty}(\boldsymbol{W},\boldsymbol{b}) \le \sum_{i=1}^{N}
\frac{1}{{\pi_{y_i}}}\log\bigg(\sum_{j=1}^{\mathcal{C}}\mathrm{E}_{\tilde{{\boldsymbol{h}}}_{i}}\big[e^{\Delta \boldsymbol{w}_{j,y_i}\tilde{\boldsymbol{h}}_{i}+\Delta b_{j,y_i}}\big]\bigg), \label{eq2}
\end{aligned}$}
\end{equation}
where $\Delta \boldsymbol{w}_{j,y_{i}}=\boldsymbol{w}_{j}^{T}-\boldsymbol{w}_{y_{i}}^{T}$ and $\Delta b_{j,y_{i}}=b_{j}-b_{y_{i}}$. 
As $\tilde{\boldsymbol{h}}_{i}$ is a Gaussian random variable adhering to 
$\mathcal{N}({\boldsymbol{{h}}}_{i}+\boldsymbol{\delta}_{i},\alpha{\boldsymbol{\Sigma}}_{y_{i}})$, 
we know that $\Delta \boldsymbol{w}_{j,y_{i}}\tilde{\boldsymbol{h}}_{i}+\Delta {b}_{j,y_{i}}\!\sim\!\mathcal{N}\big(\Delta \boldsymbol{w}_{j,y_{i}}\left({\boldsymbol{h}}_{i}+\boldsymbol{\delta}_{i}\right)+\Delta {b}_{j,y_{i}}, \alpha \Delta \boldsymbol{w}_{j,y_{i}} {\boldsymbol{\Sigma}}_{y_i}\Delta \boldsymbol{w}_{j,y_{i}}^{T}\big)$.
Subsequently, utilizing the moment-generating function,
\begin{equation}
\mathrm{E}\left[e^{t X}\right]=e^{t \mu+\frac{1}{2} \sigma^2 t^2}, \quad X \sim \mathcal{N}(\mu, \sigma^2),
\end{equation}
the upper bound of Eq.~(\ref{eq2}) can be represented as
\begin{equation}
\begin{aligned}
\mathcal{L}^{\mathcal{M}\rightarrow\infty}(\boldsymbol{W},\boldsymbol{b}) \le \sum_{i=1}^{N}-\frac{1}{\pi_{y_i}}\log \frac{e^{\mathcal{Z}_{i}^{y_i}}}{\sum_{j=1}^{\mathcal{C}}e^{\mathcal{Z}_{i}^{j}}}, 
\end{aligned}
\label{eq3}
\end{equation}
where $\mathcal{Z}_{i}^{j} = \boldsymbol{w}_{j}(\boldsymbol{h}_{i}+\boldsymbol{\delta}_{i})+b_{j}+\alpha\rho_{i}^{j}$ and $\rho_{i}^{j}\!=\!\frac{1}{2}\Delta \boldsymbol{w}_{j,y_{i}} {\boldsymbol{\Sigma}}_{y_i}\Delta \boldsymbol{w}_{j,y_{i}}^{T}$.
\begin{table}[t]
\centering
\small
\renewcommand\arraystretch{1.4}{
\begin{tabular}{p{0.7cm}<{\centering}p{4cm}<{\centering}p{2.5cm}<{\centering}}
\toprule
\textbf{Name} & \textbf{Term}                                                                                                                                                & \textbf{Function}           \\ 
\hline
$\mathcal{G}$ & $\sum_{i=1}^{N}\sum_{j\neq y_i}{q}_{i,j}\rho_{i}^{j}$                                                  & Generalization \\ 
$\mathcal{R}$ & $\sum_{i=1}^{N}\sum_{j\neq y_i}{q}_{i,j}\Delta\boldsymbol{w}_{j,y_i}\boldsymbol{\delta}_{i}$ & Robustness \\ 
$\mathscr{F}$ & $\sum_{i=1}^{N}\sum_{j\neq y_i}{q}_{i,j}\log(\pi_{j}/\pi_{y_i})$                                                                                                              & Inter-class fairness    \\
\bottomrule
\end{tabular}
}
\caption{The regularization terms incorporated in the IADA loss and their corresponding functions. $\boldsymbol{q}_{i}$ denotes the Softmax output linked to sample $\boldsymbol{x}_{i}$. The derivation process is presented in the Appendix.}
\label{tableregu}
\end{table}

Drawing inspiration from the Logit Adjustment (LA) approach~\cite{r32}, we introduce its logit adjustment term in place of the class-wise weight $1/\pi_{y_i}$, providing a more effective solution for imbalanced class distributions. Consequently, the final IADA loss is formulated as follows:
\begin{equation}
\begin{aligned}
\mathcal{L}^{{IADA}}(\boldsymbol{W},\boldsymbol{b}) \coloneq \sum_{i=1}^{N}-\log \frac{e^{{\tilde{\mathcal{Z}}}_{i}^{y_i}}}{\sum_{j=1}^{\mathcal{C}}e^{{\tilde{\mathcal{Z}}}_{i}^{j}}}, 
\end{aligned}
\label{eq4}
\end{equation}
where ${\tilde{\mathcal{Z}}}_{i}^{j} = \boldsymbol{w}_{j}(\boldsymbol{h}_{i}+\boldsymbol{\delta}_{i})+b_{j}+\alpha\mathcal{\rho}_{i}^{j}+\beta\log\pi_{j}$. The symbols $\alpha$ and $\beta$ serve as two hyperparameters in the IADA loss, frequently set to 0.5 and 1, respectively, in practical applications. 
The IADA loss, essentially a logit-adjusted variant of the CE loss, operates as a surrogacy for our proposed adversarial data augmentation strategy. 
Therefore, instead of explicitly executing the augmentation process, we can directly optimize this loss, thereby improving efficiency.

We further explain the IADA loss, defined in Eq.~(\ref{eq4}), from a regularization perspective using the Taylor expansion. 
This process reveals three regularization terms stemming from our adversarial data augmentation strategy, as outlined in Table~\ref{tableregu}. A detailed analysis of these terms is presented in Section~III of the Appendix. Through our analysis, the first term $\mathcal{G}$ diminishes the mapped variances of deep features within each class, thereby enhancing intra-class compactness and improving the generalization ability of models. The second term $\mathcal{R}$ reinforces model robustness by increasing the cosine similarity between the classification boundary and the gradient vectors of adversarial samples. Furthermore, the third term $\mathscr{F}$ promotes fairness among classes by favoring less-represented categories. Collectively, the regularization terms incorporated in the IADA loss significantly contribute to strengthening the generalization, robustness, and inter-class fairness of DNNs, as depicted in Fig.~\ref{frametu}~(Box~2). 

\section{Optimization Using IADA Loss}
\label{frame}


Due to the necessity of pre-determining the value of $\epsilon_{i}$ in $\boldsymbol{\delta}_{i}$, which governs the perturbation strategies, when utilizing the IADA loss, we construct a meta-learning-based framework named Meta-IADA. This framework is designed to optimize classifiers using the IADA loss. As illustrated in 
Fig.~\ref{frametu} (Box~3), Meta-IADA comprises four main components: the classifier, the characteristics extraction module, the perturbation network, and the meta-learning-based learning strategy.

Considering that determining perturbation strategies for samples involves factors including their learning difficulty, class distribution, and noise levels~\cite{Zhou_Yang_Wu_2023,xu2021robust}, we extract fifteen training characteristics (e.g., sample loss and margin) from the classifier to encompass these aspects. All characteristics are comprehensively introduced in Section~IV of the Appendix. These extracted characteristics then serve as inputs to the perturbation network, assisting in computing the values of $\epsilon_{i}$. Within our framework, the perturbation network employs a two-layer MLP\footnote{A comparative analysis of alternative architectures for the perturbation network is detailed in Section~V of the Appendix.}, known theoretically as a universal approximator for nearly any continuous function. Its output passes through a Tanh function to constrain the range of $\epsilon_{i}$ within $(-1, 1)$.



\begin{algorithm}[t]
\caption{Algorithm of Meta-IADA}
\label{alg1}
\LinesNumbered 
\KwIn{Training data $\mathcal{D}^{tr}$, metadata ${\mathcal{D}}^{meta}$, batch sizes $n$ and $m$, ending iterations $\mathcal{T}_{1}$ and $\mathcal{T}_{2}$. }
\KwOut{Learned classifier parameter ${\boldsymbol{\Phi}}$.
}
{Initialize classifier parameter ${\boldsymbol{\Phi}}^{1}$ and perturbation network parameter $\boldsymbol{\Omega}^{1}$}\;
\For {$t \le \mathcal{T}_{1}$}{
    Sample 
    $ \{(\boldsymbol{x}_{i},{y}_{i})\}_{i=1}^{n}$ from $\mathcal{D}^{tr}$\;
    Update 
    ${\boldsymbol{\Phi}}^{t+1}\leftarrow{\boldsymbol{\Phi}}^{t}-\eta_{1} \frac{1}{n} \sum\nolimits_{i=1}^{n}\nabla_{{{\boldsymbol{\Phi}}}}\ell^{{CE}}_{i}$\;
}  
\For {$\mathcal{T}_{1}<t \le \mathcal{T}_{2}$}{
    Sample 
    $ \{(\boldsymbol{x}_{i},{y}_{i})\}_{i=1}^{n}$ from $\mathcal{D}^{tr}$\;
    Sample 
    $\{(\boldsymbol{x}^{meta}_{i},{y}^{meta}_{i})\}_{i=1}^{m}$ form $\mathcal{D}^{meta}$\;

    Obtain current covariance matrices ${\boldsymbol{\Sigma}}$\;
    Formulate 
    $\overline{{\boldsymbol{\Phi}}}^{t+1}$ by Eq.~(\ref{meta1})\;
    Update 
    $\boldsymbol{\Omega}^{t+1}$ by Eq.~(\ref{meta2})\;
    Update 
   ${\boldsymbol{\Sigma}}^{t+1}$ by Eq.~(\ref{meta3})\;
    Update 
    ${\boldsymbol{\Phi}}^{t+1}$ by Eq.~(\ref{meta5})\;

}  
\end{algorithm}


Meta-IADA employs a meta-learning-based strategy to iteratively update the classifier and the perturbation network. This involves leveraging a high-quality (clean and balanced) yet small meta dataset, $\mathcal{D}^{{meta}}\!=\!\{(\boldsymbol{x}_{i}^{{meta}},y_{i}^{{meta}})\}_{i=1}^{M}$. Additionally, to mitigate the accuracy compromise within ${\boldsymbol{\Sigma}}_{y_i}$ for minor classes due to the constraints of limited training data, their values are further updated on the metadata. Let $\boldsymbol{\Omega}$ represent the parameters of the perturbation network. The optimization process within Meta-IADA is detailed as follows:


Initially, the parameters of the classifier, $\boldsymbol{\Phi}$, are updated using stochastic gradient descent (SGD) on a mini-batch of training samples $\{(\boldsymbol{x}_{i},{y}_{i})\}_{i=1}^{n}$ with the following objective:
\begin{equation}
\resizebox{1\hsize}{!}{$
\begin{aligned}
\overline{{\boldsymbol{\Phi}}}^{t+1}\leftarrow{\boldsymbol{\Phi}}^{t}-\eta_{1} \frac{1}{n} \sum\nolimits_{i=1}^{n} \nabla_{{{\boldsymbol{\Phi}}}}\ell^{{IADA}}\big({\mathcal{F}}(\boldsymbol{x}_{i};{\boldsymbol{\Phi}}^{t}),y_{i}; {\boldsymbol{\Sigma}}^{t}_{y_i},\epsilon({\boldsymbol{f}}_{i}^{t},{\boldsymbol{\Omega}}^{t})\big),
\end{aligned}$}
\label{meta1}
\end{equation}
where 
$\eta_{1}$ is the step size. Additionally, $\boldsymbol{f}_{i}^{t}$ denotes the concatenated vector containing the extracted training characteristics for $\boldsymbol{x}_{i}$ at the $t$th iteration. Subsequently, leveraging the optimized $\overline{\boldsymbol{\Phi}}^{t+1}$, the parameter update in the perturbation network, $\boldsymbol{\Omega}$, entails utilizing a mini-batch of metadata $\{(\boldsymbol{x}_{i}^{{meta}},{y}_{i}^{{meta}})\}_{i=1}^{m}$:
\begin{equation}
\resizebox{1\hsize}{!}{$
\begin{aligned}
\boldsymbol{\Omega}^{t+1}&\leftarrow \boldsymbol{\Omega}^{t}-\eta_{2} \frac{1}{m} \sum\nolimits_{i=1}^{m} \nabla_{\boldsymbol{\Omega}}\ell^{{CE}}\big({\mathcal{F}}(\boldsymbol{x}_{i}^{{meta}};\overline{\boldsymbol{\Phi}}^{t+1}),y_{i}^{{meta}}\big),
\end{aligned}$}
\label{meta2}
\end{equation}
where 
$\eta_{2}$ denotes the step size. Simultaneously, the optimization of covariance matrices is conducted using the metadata, outlined as follows:
\begin{equation}
\resizebox{1\hsize}{!}{$
\begin{aligned}
    {\boldsymbol{\Sigma}}^{t+1} & \leftarrow{\boldsymbol{\Sigma}}^{t}-\eta_{2} \frac{1}{m} \sum\nolimits_{i=1}^{m} \nabla_{\boldsymbol{\Sigma}}\ell^{{CE}}\big({\mathcal{F}}(\boldsymbol{x}_{i}^{{meta}};\overline{\boldsymbol{\Phi}}^{t+1}),y_{i}^{{meta}}\big).
\end{aligned}$}
\label{meta3}
\end{equation}
Finally, leveraging the computed perturbations and updated covariance matrices, we proceed to update the parameters $\boldsymbol{\Phi}$ of the classifier in the following manner:
\begin{equation}
\resizebox{1\hsize}{!}{$
\begin{aligned}
{\boldsymbol{\Phi}}^{t+1}&\leftarrow{\boldsymbol{\Phi}}^{t}-\eta_{1} \frac{1}{n} \sum\nolimits_{i=1}^{n}\nabla_{{{\boldsymbol{\Phi}}}}\ell^{{IADA}}\big({\mathcal{F}}(\boldsymbol{x}_{i};\boldsymbol{\Phi}^{t}),y_{i}; {\boldsymbol{\Sigma}}^{t+1}_{y_i},\epsilon({\boldsymbol{f}}_{i}^{t},{\boldsymbol{\Omega}}^{t+1})\big).
\end{aligned}$}
\label{meta5}
\end{equation}
Similar to MetaAug~\cite{r9}, to acquire better-generalized representations, the classifier is initially trained using vanilla CE loss, followed by trained with Meta-IADA. The algorithm for Meta-IADA is delineated in Algorithm~\ref{alg1}.

\begin{table}[t]
\centering
\resizebox{0.48\textwidth}{!}{
\begin{tabular}{l|cc|cc}
\toprule
{\textbf{Dataset}}                               & \multicolumn{2}{c|}{\textbf{CIFAR10}}      & \multicolumn{2}{c}{\textbf{CIFAR100}}                                                   \\  

\textbf{Imbalance ratio}& \multicolumn{1}{c}{\textbf{100:1}}                          & \textbf{10:1}    & \multicolumn{1}{c}{\textbf{100:1}}                          & \textbf{10:1}                        \\   \midrule
Class-Balanced CE$^{\dagger}$~\cite{r42}                 & \multicolumn{1}{c}{72.68}                        & 86.90       & \multicolumn{1}{c}{38.77}                        & 57.57                   \\
Class-Balanced Focal$^{\dagger}$~\cite{r42}              & \multicolumn{1}{c}{74.57}                        & 87.48     & \multicolumn{1}{c}{39.60}                        & 57.99                     \\
LDAM-DRW$^{\dagger}$~\cite{r30}                               & \multicolumn{1}{c}{78.12}                        & 88.37      & \multicolumn{1}{c}{42.89}                        & 58.78                           \\
{Meta-Weight-Net$^{\dagger}$~\cite{r46}} & \multicolumn{1}{c}{{73.57}} & {87.55} & \multicolumn{1}{c}{{41.61}} & {58.91}  \\
De-confound-TDE$^{\ast}$~\cite{r45}                        & \multicolumn{1}{c}{{80.60}}                        & 88.50  & \multicolumn{1}{c}{44.10}                        & 59.60                                \\
LA~\cite{r32}                                      & \multicolumn{1}{c}{77.67}                        & 88.93      & \multicolumn{1}{c}{43.89}                        & 58.34                            \\
MiSLAS$^{\ast}$~\cite{zhong2021improving}                        & \multicolumn{1}{c}{{82.10}}                        &{{90.00}}     & \multicolumn{1}{c}{{47.00}}                        & {{63.20}} \\
LADE~\cite{hong2021disentangling}                        & \multicolumn{1}{c}{{81.17}}                        & 89.15     & \multicolumn{1}{c}{{45.42}}                        & {61.69} \\
{MetaSAug$^{\dagger}$~\cite{r9}}        & \multicolumn{1}{c}{{{80.54}}} & {{89.44}} & \multicolumn{1}{c}{{{46.87}}} & {{61.73}} \\ 
LPL$^{\ast}$~\cite{r33}                                    & {77.95}                        & {89.41}      & {{44.25}}                        & {60.97}                       \\
RISDA$^{\ast}$~\cite{r79}                       & \multicolumn{1}{c}{{79.89}}                        & 89.36     & \multicolumn{1}{c}{{50.16}}                        & {62.38} 
\\
LDAM-DRW-SAFA$^{\ast}$~\cite{hong2022safa}                        & \multicolumn{1}{c}{{80.48}}                        & 88.94     & \multicolumn{1}{c}{{46.04}}                        & {59.11} \\
BKD$^{\ast}$~\cite{zhang2023balanced}                        & \multicolumn{1}{c}{{82.50}}                        & 89.50     & \multicolumn{1}{c}{{46.50}}                        & {62.00} \\
\rowcolor{mytablecolor2}
{{Meta-IADA (Ours)}}      & \multicolumn{1}{c}{{{\textbf{84.01}}}} & {{\textbf{91.73}}}   & \multicolumn{1}{c}{{{\textbf{52.18}}}} & {{\textbf{64.72}}}  \\ 
\bottomrule
\end{tabular}}
\caption{Accuracy (\%) comparison on CIFAR-LT benchmark. The best results are highlighted in bold. ``$\ast$" represents the results in the original paper, while ``$\dagger$" denotes the results in~\protect\cite{r9}.}
\label{table2}
\end{table}

\begin{figure}[bp] 
\centering
\includegraphics[width=0.48\textwidth]{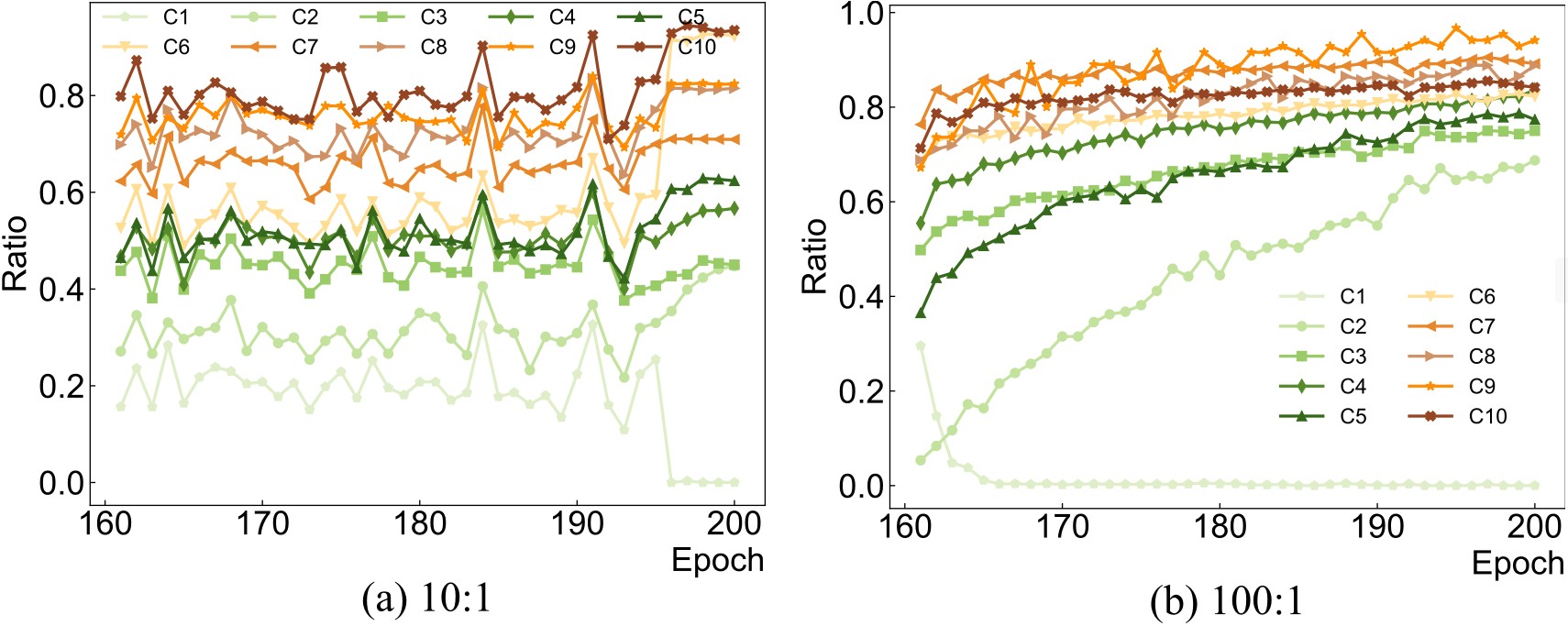}
\caption{Ratio of adversarial samples in each class during the last forty epochs on CIFAR10 under imbalance ratios of 10:1 and 100:1. From ``C1" to ``C10", the class proportions progressively decrease.
}
\label{imb11}
\end{figure}

\section{Experiments}
We experiment across four typical biased learning scenarios, including LT learning, GLT learning, noisy label learning, and subpopulation shift learning, involving image and text datasets. The excluded settings and results (including those on standard datasets) are detailed in the Appendix.

\begin{table}[t]
\small
\centering
\begin{tabular}{p{6.1cm}|p{1.4cm}<{\centering}}
\toprule
\textbf{Method}               & \textbf{Accuracy$\uparrow$} \\ \midrule
Class-Balanced CE$^{\dagger}$~\cite{r42}    &  66.43     \\
Class-Balanced Focal$^{\dagger}$~\cite{r42} & 61.12       \\
LDAM-DRW$^{\dagger}$~\cite{r30}            &  {68.00}     \\
BBN$^{\dagger}$~\cite{r69}        & 66.29      \\
Meta-Class-Weight$^{\dagger}$~\cite{meta_class_weight}    &  67.55    \\
MiSLAS$^{\ast}$~\cite{zhong2021improving}         &  {71.60}    \\
LADE$^{\ast}$~\cite{hong2021disentangling}   & {70.00}       \\
MetaSAug$^{\dagger}$~\cite{r9}             &  {68.75}     \\ 
LDAM-DRS-SAFA$^{\ast}$~\cite{hong2022safa}       &  {69.78}     \\
RISDA$^{\ast}$~\cite{r79}       &  {69.15}  \\
BKD$^{\ast}$~\cite{zhang2023balanced}       &  {71.20}   \\ \rowcolor{mytablecolor2}
Meta-IADA (Ours)            &    \textbf{72.55}   \\\bottomrule
\end{tabular}
\caption{Accuracy (\%) comparison on iNat 2018 dataset.}
\label{ina_result}
\end{table}


\subsection{Long-Tail Learning}
\label{imbalance}
Four LT image classification benchmarks, CIFAR-LT~\cite{r42}, ImageNet-LT~\cite{r81}, Places-LT~\cite{r81}, and iNaturalist (iNat) 2018~\cite{meta_class_weight}, are evaluated. Additionally, two imbalanced 
text classification datasets 
are included. Due to space limitations, we only present experiments for CIFAR-LT and iNat in the main text. 

\paragraph{Experiments on CIFAR-LT Datasets.}
We employ the ResNet-32 model~\cite{r36} with an initial learning rate of 0.1. The training employs the SGD optimizer with a momentum of 0.9 and a weight decay of $5\!\times\!10^{-4}$ on a single GPU, spanning 200 epochs. The learning rate is decayed by 0.01 at the $160$th and $180$th epochs. Additionally, the perturbation network is optimized using Adam, with an initial learning rate of $1\!\times\!10^{-3}$. To construct metadata, we randomly select ten images per class from the validation data. For the hyperparameters in the IADA loss, $\alpha$ is selected from \{0.1, 0.25, 0.5, 0.75, 1\}, while keeping $\beta$ fixed at 1.

From the results reported in Table~\ref{table2}, Meta-IADA displays remarkable superiority over other LT baselines, underscoring its effectiveness in managing imbalanced class distributions. Additionally, it outperforms previous implicit augmentation methods, providing evidence for the efficacy of augmenting samples within their adversarial and anti-adversarial perturbation distributions. Furthermore, as depicted in Fig.~\ref{imb11}, Meta-IADA induces a lower proportion of adversarial samples in major classes (``C1" to ``C5"), whereas minor classes (``C6" to ``C10") exhibit a higher proportion. This manifests the model's increased attention on samples in tail classes. Comparative analyses of confusion matrices involving CE loss, MetaSAug, and Meta-IADA are presented in Section~V of the Appendix. 
The results unveil that Meta-IADA remarkably increases the accuracy of both major and minor classes. Conversely, while MetaSAug improves the performance of minor categories, it compromises the accuracy of major classes.

\paragraph{Experiments on iNat 2018 Dataset.}
The ResNet-50 model serves as the backbone classifier, pre-trained on ImageNet~\cite{r41} and iNat 2017~\cite{ina2017} datasets. The perturbation network is optimized using Adam, with an initial learning rate of $1\!\times\!10^{-4}$. Other settings follow those outlined in the MetaSAug~\cite{r9} paper. 
As reported in Table~\ref{ina_result}, Meta-IADA outperforms other comparative approaches tailored for LT learning. This suggests that, even in situations with imbalanced class distributions, employing a sample-level strategy proves more effective, as finer-grained imbalances within each class may also exist.

\begin{table}[t]
\centering
\resizebox{0.48\textwidth}{!}{
\begin{tabular}{l|cc|cc|cc}
\toprule
{\textbf{Protocol}}     & \multicolumn{2}{c|}{\textbf{CLT}}  & \multicolumn{2}{c|}{\textbf{GLT}} & \multicolumn{2}{c}{\textbf{ALT}} \\ 
\textbf{Metric}& \textbf{Acc.$\uparrow$}        & \textbf{Prec.$\uparrow$}      & \textbf{Acc.$\uparrow$}        & \textbf{Prec.$\uparrow$}      & \textbf{Acc.$\uparrow$}        & \textbf{Prec.$\uparrow$}      \\ \midrule
CE loss$^{\ddag}$    &42.52  &47.92   & 34.75       & 40.65      & 41.73       & 41.74      \\
MixUp$^{\ddag}$~\cite{r44}     & 38.81 & 45.41      & 31.55 & 37.44      &42.11 & 42.42     \\
LDAM$^{\ddag}$~\cite{r30}  &      {46.74} & 46.86   & {38.54}       & 39.08      & 42.66       & 41.80      \\
ISDA~\cite{r8}     & {42.66} & 44.98& {36.44}   &  37.26  & {43.34} & {43.56} \\
cRT$^{\ddag}$~\cite{r68}       &45.92 &45.34     & 37.57       & 37.51     & 41.59      & 41.43     \\
LWS$^{\ddag}$~\cite{r68}        &46.43 &45.90    & 37.94      & 38.01     & 41.70       & 41.71      \\
De-confound-TDE$^{\ddag}$~\cite{r45} & 45.70 &44.48 &37.56      & 37.00      & 41.40       & 42.36      \\
BLSoftmax$^{\ddag}$~\cite{r70} &  45.79 & 46.27   & 37.09       & 38.08      & 41.32       & 41.37      \\
BBN$^{\ddag}$~\cite{r69}     &  46.46 & 49.86     & 37.91       & 41.77     & 43.26       & 43.86      \\
RandAug$^{\ddag}$~\cite{r71}     &46.40 & {52.13}     & 38.24 & {44.74}   & {46.29} &46.32   \\
LA$^{\ddag}$~\cite{r32}&   46.53 & 45.56  & 37.80       & 37.56     & -        & -        \\
MetaSAug~\cite{r9} & {48.53} & {54.21} & {40.27} & {44.38} & {47.62} & {48.26} \\
IFL$^{\ddag}$~\cite{r67}        & 45.97 & 52.06    & 37.96       & 44.47      & 45.89       & {46.42}      \\

RISDA~\cite{r79}     & {46.31} & 51.24& {38.45}   &  42.77  & {43.65} & {43.23} \\
BKD~\cite{zhang2023balanced}      & {46.51} &     {50.15} &  {37.93}         &  {41.50}           &    {42.17}          &   {41.83}         \\
\rowcolor{mytablecolor2}
Meta-IADA (Ours)      & \textbf{53.45} &     \textbf{58.05} &  \textbf{44.36}         &   \textbf{50.07}      &    \textbf{52.54}         &   \textbf{53.23}         \\ 
\bottomrule
\end{tabular}}
\caption{Accuracy and precision (\%) of CLT, GLT, and ALT protocols on the ImageNet-GLT benchmark. $\ddag$ indicates the results reported in \protect\cite{r67}.}
\label{glt1}
\end{table}

\begin{figure}[bp] 
\centering
\includegraphics[width=0.48\textwidth]{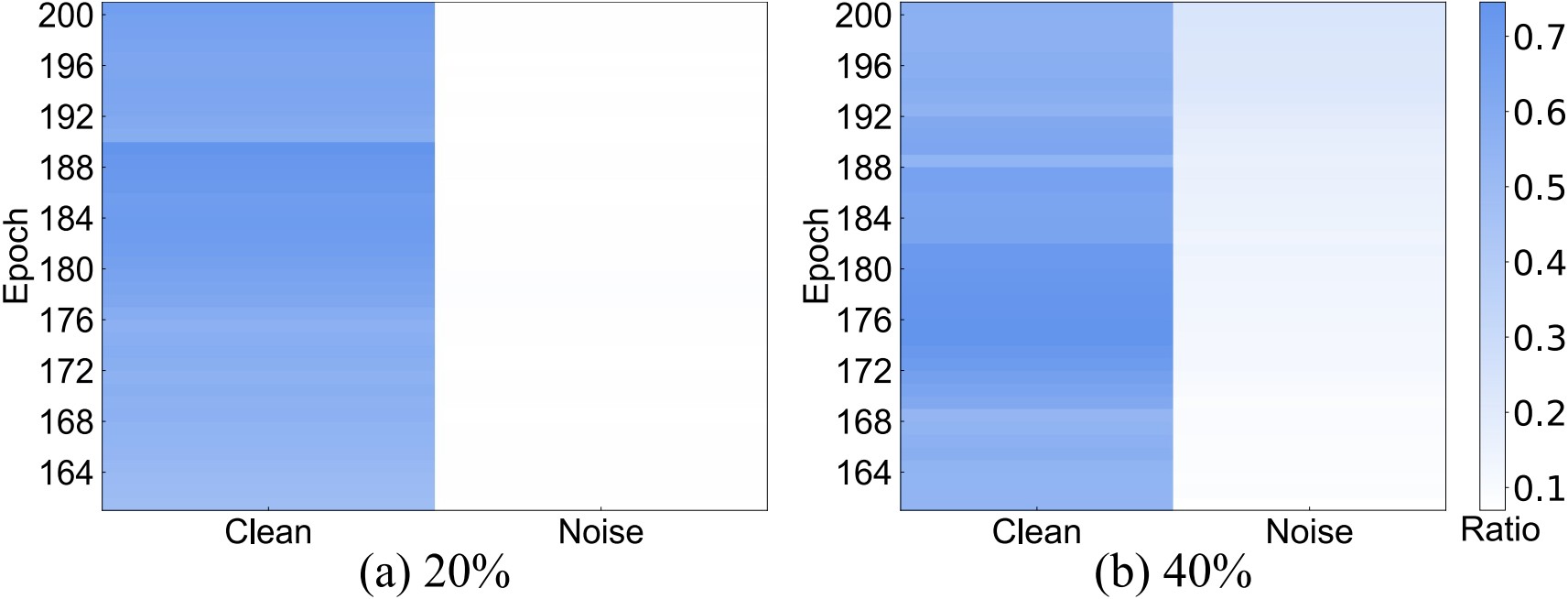}
\caption{Ratio of adversarial samples for noisy and clean samples in the final forty epochs on CIFAR10 with 20\% and 40\% flip noise. 
}
\label{fig99}
\end{figure}

\subsection{Generalized Long-Tail Learning}
GLT learning considers both long-tailed class and attribute distributions in the training data. We employ two GLT benchmarks, ImageNet-GLT, and MSCOCO-GLT~\cite{r67}. Each benchmark comprises three protocols, CLT, ALT, and GLT, showcasing variations in class distribution, attribute distribution, and combinations of both between training and testing datasets. The ResNeXt-50~\cite{r37} model is utilized as the backbone network. We train models with a batch size of 256 and an initial learning rate of 0.1, using SGD with a weight decay of $5\!\times\!10^{-4}$ and a momentum of 0.9. The perturbation network is optimized using the Adam optimizer, initialized with a learning rate of $1\!\times\!10^{-3}$. The metadata utilized in this experiment is balanced in both classes and attributes. We adopt the construction method of Tang et al.~\shortcite{r67}, which involves clustering images in each class into six groups using KMeans with a pre-trained ResNet-50 model. Two images are then randomly selected from each group and class within the validation data.


The comparative results for ImageNet-GLT are outlined in Table~\ref{glt1}, while those for MSCOCO-GLT are presented in the Appendix. Meta-IADA demonstrates substantial enhancements in accuracy and precision across all three protocols, emphasizing its ability to address distributional skewness in three scenarios, including attribute imbalance, class imbalance, and their combination. The efficacy of Meta-IADA stems from its ability to apply adversarial augmentation to samples from tail classes and those with rare attributes. This significantly magnifies the impact of these samples on model training, resulting in improved prediction accuracy. 
However, methods specifically designed for LT learning demonstrate subpar performance on the ALT protocol, primarily due to their class-level characteristics.

\begin{table}[t]
\centering
\footnotesize
\resizebox{0.48\textwidth}{!}{
\begin{tabular}{l|cc|cc}
\toprule
{\textbf{Dataset}}           & \multicolumn{2}{c|}{\textbf{CIFAR10}} & \multicolumn{2}{c}{\textbf{CIFAR100}}                                                    \\ 
\textbf{Noise ratio}    & \multicolumn{1}{c}{\textbf{20\%}}                         & \textbf{40\%}        & \multicolumn{1}{c}{\textbf{20\%}}                         & \textbf{40\%}                   \\ \midrule
CE loss & \multicolumn{1}{c}{76.85}                        & 70.78    & \multicolumn{1}{c}{50.90}                        & 43.02                      \\
D2L~\cite{r49}               & \multicolumn{1}{c}{87.64}                        & 83.90                 & \multicolumn{1}{c}{63.39}                        & 51.85             \\
Co-teaching~\cite{r48}       & \multicolumn{1}{c}{82.85}                        & 75.43    & \multicolumn{1}{c}{54.19}                        & 44.92                      \\ 
GLC~\cite{r53}               & \multicolumn{1}{c}{89.77}                        & {88.93}    & \multicolumn{1}{c}{63.15}                        & {62.24}                \\
MentorNet~\cite{r51}         & \multicolumn{1}{c}{86.41}                        & 81.78   & \multicolumn{1}{c}{62.00}                        & 52.71                          \\
L2RW~\cite{r52}              & \multicolumn{1}{c}{87.88}                        & 85.70   & \multicolumn{1}{c}{57.51}                        & 51.00                       \\
{DMI}~\cite{zaoshengl}               & \multicolumn{1}{c}{88.43}                        & 84.00         & \multicolumn{1}{c}{58.87}                        & 42.95                           \\


Meta-Weight-Net~\cite{r46}   & \multicolumn{1}{c}{{90.35}}                        & 87.65      & \multicolumn{1}{c}{64.31}                        & 58.67                        \\
APL~\cite{r50}               & \multicolumn{1}{c}{87.45}                        & 81.08   & \multicolumn{1}{c}{59.86}                        & 53.31                      \\
JoCoR~\cite{zaosheng2}        & \multicolumn{1}{c}{{90.78}}                        & 83.67                  & \multicolumn{1}{c}{{65.21}}                        & 46.44        \\
MLC~\cite{zheng2021meta}              & \multicolumn{1}{c}{{91.55}}                        & {89.53}  & \multicolumn{1}{c}{{66.32}}                        & {62.29}                        \\
MFRW-MES~\cite{ricci2023meta}              & \multicolumn{1}{c}{{91.45}}                        & {90.72}  & \multicolumn{1}{c}{{65.27}}                        & {62.35}                        \\

\rowcolor{mytablecolor2}
Meta-IADA (Ours)       & \multicolumn{1}{c}{{\textbf{93.44}}} & {\textbf{91.99}} & \multicolumn{1}{c}{{\textbf{69.16}}} & {\textbf{64.37}} \\ 
\bottomrule
\end{tabular}}
\caption{Accuracy (\%) comparison on CIFAR datasets with 20\% and 40\% flip noise. 
}
\label{flip1}
\end{table}

\begin{table}[bp]
\centering
\resizebox{0.48\textwidth}{!}{
\begin{tabular}{l|cc|cc|cc|cc}
\toprule
         {\textbf{Dataset}} & \multicolumn{2}{c|}{\textbf{CelebA}} & \multicolumn{2}{c}{\textbf{CMNIST}} & \multicolumn{2}{c|}{\textbf{Waterbirds}} & \multicolumn{2}{c}{\textbf{CivilComments}}\\ 
         \textbf{Metric} & \textbf{Avg.$\uparrow$}         & \textbf{Worst$\uparrow$}        & \textbf{Avg.$\uparrow$}          & \textbf{Worst$\uparrow$}     & \textbf{Avg.$\uparrow$}          & \textbf{Worst$\uparrow$}  & \textbf{Avg.$\uparrow$}          & \textbf{Worst$\uparrow$}           \\ \midrule
CORAL$^{\dotplus}$~\cite{DORAL}     & 93.8       & 76.9      & 71.8       & 69.5   & 90.3           & 79.8         & 88.7            & 65.6          \\
IRM$^{\dotplus}$~\cite{r58}       & {94.0}       & 77.8       & 72.1     & 70.3     & 87.5          & 75.6         & 88.8            & 66.3       \\
GroupDRO$^{\dotplus}$~\cite{r54}  & 92.1       & 87.2       & 72.3      & 68.6      & 91.8           & {90.6}         & 89.9           & 70.0  \\
DomainMix$^{\dotplus}$~\cite{r65} & 93.4      & 65.6       & 51.4       & 48.0      & 76.4           & 53.0         & {90.9}           & 63.6         \\
IB-IRM$^{\dotplus}$~\cite{r59}    & 93.6       & 85.0      & 72.2      & 70.7     & 88.5           & 76.5        & 89.1            & 65.3        \\
V-REx$^{\dotplus}$~\cite{r60}     & 92.2       & 86.7       & 71.7       & 70.2     & 88.0           & 73.6        & 90.2           & 64.9       \\
UW$^{\dotplus}$~\cite{r57}        & 92.9       & 83.3       & 72.2       & 66.0     & \textbf{95.1}          & 88.0         & 89.8           & 69.2   \\
Fish$^{\dotplus}$~\cite{r66}      & 93.1       & 61.2       & 46.9       & 35.6      & 85.6           & 64.0         & 89.8            & 71.1          \\
LISA$^{\dotplus}$~\cite{r57}      & 92.4      & {89.3}       & {74.0}       & {73.3}  & 91.8           & 89.2         & 89.2            & {72.6}        \\ \rowcolor{mytablecolor2}
Meta-IADA (Ours)      & \textbf{94.5}       & \textbf{91.2}       & \textbf{78.0}       & \textbf{75.9}    & {94.5}           & \textbf{92.5}         & \textbf{92.1}            & \textbf{74.8}      \\ 
\bottomrule
\end{tabular}}
\caption{Comparison of the average and worst-group accuracy (\%) on four subpopulation shift datasets. ${\dotplus}$ indicates the results reported in \protect\cite{r57}.}
\label{table4}
\end{table}

\subsection{Noisy Label Learning}
\label{noise}
We examine two types of label corruptions: uniform and pair-flip noise~\cite{r46}, using CIFAR~\cite{krizhevsky2009learning} datasets. The Wide ResNet-28-10 (WRN-28-10)~\cite{r38} and ResNet-32 models are utilized for uniform and flip noises, respectively.
1,000 images with clean labels are selected from the validation set to compile the metadata. The ResNet settings match those used for CIFAR-LT. WRN-28-10 is trained using SGD with an initial learning rate of 0.1, a momentum of 0.9, and a weight decay of $5\!\times\!10^{-4}$; the learning rate is decayed by 0.1 at the $50$th and $55$th epochs during the total 60 epochs. The Adam optimizer, initialized with a learning rate of $1\!\times\!10^{-3}$, is utilized for optimizing the perturbation network.


Table~\ref{flip1} presents the comparison results for flip noise, while those for uniform noise are detailed in the Appendix. Meta-IADA consistently achieves SOTA performance when compared to all other methods, surpassing the highest accuracy among comparative methods by an average of 2\%.
These outcomes underscore its capacity to bolster DNNs' robustness against noise. An analysis of Fig.~\ref{fig99} indicates that nearly all noisy samples undergo anti-adversarial augmentations in Meta-IADA, effectively mitigating the negative impact of noisy samples on the overall performance of DNNs.

\begin{figure}[t] 
\centering
\includegraphics[width=0.48\textwidth]{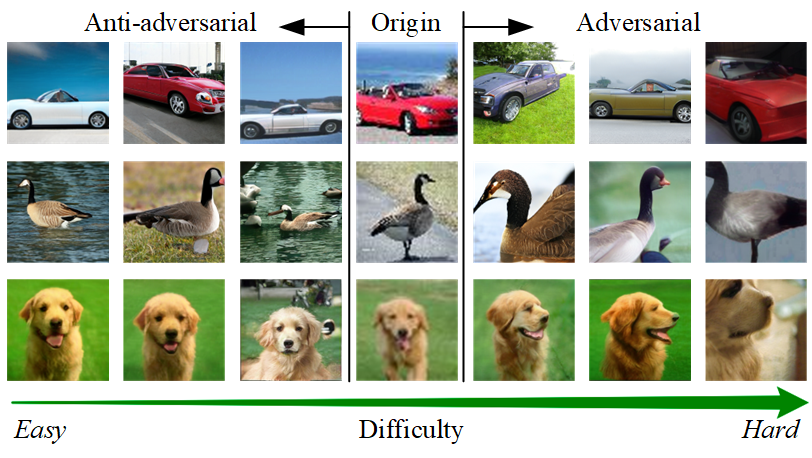}
\caption{Visualization of instances corresponding to deep features augmented by Meta-IADA.
}
\label{fig8}
\end{figure}



\begin{table}[bp]
\centering
\small
\begin{tabular}{p{0.8cm}<{\centering}p{0.8cm}<{\centering}p{0.8cm}<{\centering}|p{0.8cm}<{\centering}p{0.8cm}<{\centering}|p{0.8cm}<{\centering}p{0.8cm}<{\centering}}
\toprule
\multicolumn{3}{c|}{\textbf{Regularization terms}} & \multicolumn{2}{c|}{\textbf{CIFAR10}} & \multicolumn{2}{c}{\textbf{CIFAR100}} \\
                 {$\mathcal{G}$} & {$\mathcal{R}$}    &  {$\mathscr{F}$}                & \textbf{10:1}         & \textbf{100:1}        & \textbf{10:1}         & \textbf{100:1}         \\ \midrule
                  \checkmark &        \checkmark             &    \checkmark                 &  \textbf{8.17}             &   \textbf{15.99}           &    \textbf{35.28}          & \textbf{47.82}              \\ 
                 
                \text{\ding{55}}     &               \checkmark   &    \checkmark                 &      {9.48}        &   18.20           &     {36.95}         &      51.01         \\ 
              \checkmark     &     \text{\ding{55}}                &       \checkmark            &     10.06         &  19.11            &   37.62           &       52.47        \\ \checkmark  &                   \checkmark &  \text{\ding{55}}                  &    {9.77}          &  {17.36}            &   {37.18}           &      {49.87}         \\ \bottomrule
\end{tabular}
\caption{Results of ablation studies using the ResNet-32 model on CIFAR-LT benchmark. Top-1 error rates are reported.
}
\label{abatable}
\end{table}


\subsection{Subpopulation Shift Learning}

We conduct evaluations on four binary classification datasets characterized by subpopulation shifts: Colored MNIST (CMNIST)~\cite{r57}, Waterbirds~\cite{r54}, CelebA~\cite{r55}, and CivilComments~\cite{r56}. For three image datasets (i.e., CMNIST, Waterbirds, and CelebA), we utilize ResNet-50~\cite{r36} as the backbone network, while for the text dataset, CivilComments, we employ DistilBert~\cite{r39}. To provide a comprehensive assessment, we report both average and worst-group accuracy. Detailed experimental settings and dataset introductions are available in Section~V of the Appendix.

Based on the results in Table~\ref{table4}, Meta-IADA achieves the highest worst-group accuracy across the four datasets. This emphasizes its effectiveness in enhancing performance for underrepresented groups, such as samples in the landbird class with a water background. Typically, these samples benefit from 
adversarial augmentation, enhancing their influence on model training. Furthermore, except for the Waterbirds dataset, Meta-IADA surpasses other methods in terms of average accuracy. These findings illustrate Meta-IADA's capability to bolster model resilience against subpopulation shifts.

\subsection{Visualization Results}

We utilize the visualization method introduced in ISDA~\cite{r8}, projecting the features augmented by Meta-IADA back into the pixel space. The results, depicted in Fig.~\ref{fig8}, highlight the diversity encapsulated within the generated adversarial and anti-adversarial samples. Additionally, samples created through adversarial and anti-adversarial augmentation commonly showcase varying levels of difficulty compared to the original samples. These modifications are attributed to transformations within the deep feature space, influenced by semantic vectors associated with attributes such as angles, colors, and backgrounds.

\subsection{Ablation and Sensitivity Studies}

\begin{figure}[t] 
\centering
\includegraphics[width=0.48\textwidth]{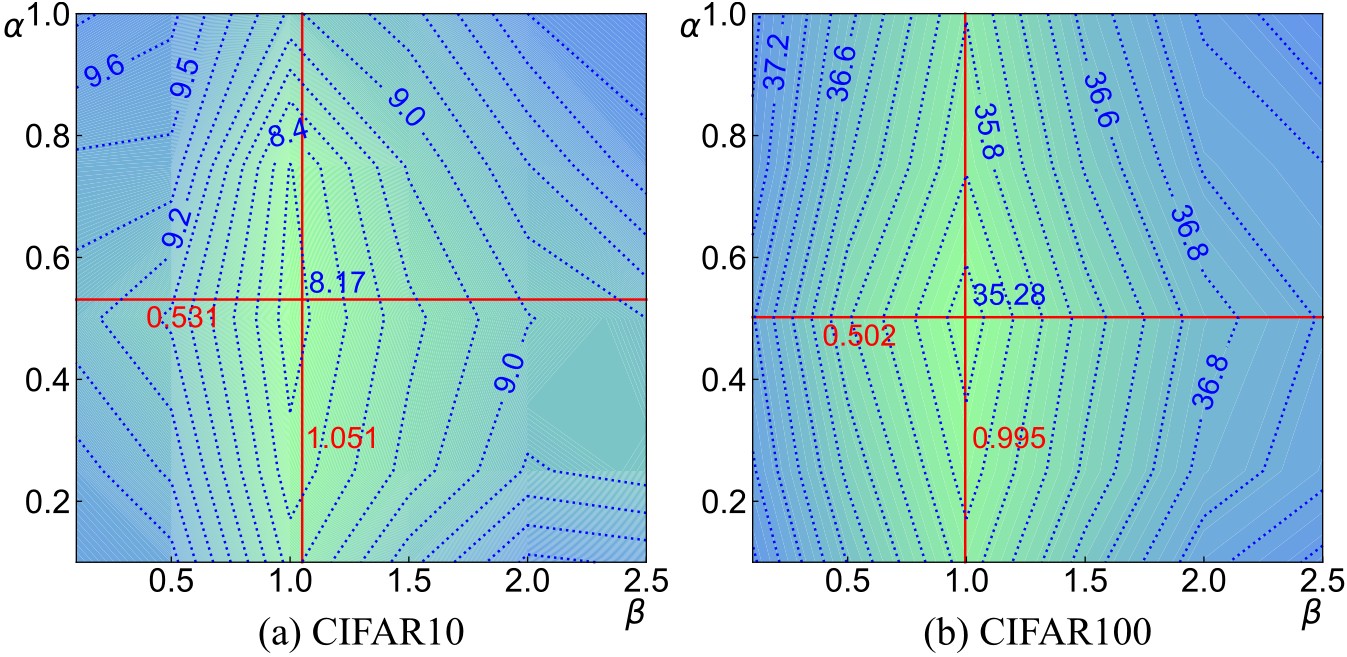}
\caption{Results of sensitivity tests on CIFAR datasets with an imbalance ratio of 10:1, using the ResNet-32 model.}
\label{aba}
\end{figure}

Ablation studies are conducted to analyze the impact of the three regularization terms within the IADA loss. The results reported in Table~\ref{abatable}, emphasize the crucial roles played by both the generalization term $\mathcal{G}$ and the robustness term $\mathcal{R}$. Additionally, in the context of an imbalanced scenario, the fairness term $\mathscr{F}$ also demonstrates its significance. Furthermore, sensitivity tests are performed on the hyperparameters $\alpha$ and $\beta$, which respectively control the effects of $\mathcal{G}$ and $\mathscr{F}$. As shown in Fig.~\ref{aba}, 
Meta-IADA achieves optimal performance when $\alpha$ is around 0.5 and $\beta$ approaches 1. Besides, the stable ranges for $\alpha$ and $\beta$ lie within $[0.25, 0.75]$ and $[0.75, 1.25]$, respectively. 
Based on these findings, we recommend setting $\alpha\!=\!0.5$ and $\beta\!=\!1$ for practical applications. 

\section{Conclusion}
This paper presents a novel adversarial data augmentation strategy to facilitate model training across diverse learning scenarios, particularly those with data biases. This strategy enriches the deep features of samples by incorporating their adversarial and anti-adversarial perturbation distributions, dynamically adjusting the learning difficulty of training samples. Subsequently, we formulate a surrogate loss for our augmentation strategy and establish a meta-learning framework to optimize classifiers using this loss. Extensive experiments are conducted across various biased learning scenarios involving different networks and datasets, showcasing the effectiveness and broad applicability of our approach.






\bibliographystyle{named}
\bibliography{ijcai24}

\end{document}